\documentclass[letterpaper]{article} 
\usepackage[preprint]{aaai2027}  
\usepackage[hyphens]{url}  
\usepackage{graphicx} 
\usepackage{natbib}  
\usepackage{caption} 
\usepackage{booktabs}
\usepackage{amsmath}
\usepackage{amssymb}
\title{ViP-Rig: Visual-Prompted Controllable Rigging}

\author{
    Zihan Qin\textsuperscript{\rm 1}\equalcontrib,
    Mingze Sun\textsuperscript{\rm 2}\equalcontrib,
    Yifan Mao\textsuperscript{\rm 1},
    Jialei Xu\textsuperscript{\rm 1},\\
    Jingfeng Guo\textsuperscript{\rm 3},
    Changrong Hu\textsuperscript{\rm 4},
    Wenbo Zhao\textsuperscript{\rm 1},
    Junjun Jiang\textsuperscript{\rm 1},
    Xianming Liu\textsuperscript{\rm 1}\corresponding
}

\affiliations{
    \textsuperscript{\rm 1}Harbin Institute of Technology\\
    \textsuperscript{\rm 2}Tsinghua University\\
    \textsuperscript{\rm 3}South China University of Technology\\
    \textsuperscript{\rm 4}University of Science and Technology of China
}

\begin{document}

\maketitle

\begin{abstract}
Rigging is inherently task-dependent because the same mesh may require different skeletons and deformation behaviors across animation tasks. In practice, artists often inspect an initial rig and repeatedly edit its skeletal structure and deformation behavior to meet specific animation requirements. Existing automatic methods primarily generate a plausible rig from geometry, offering limited explicit control over the resulting skeleton and deformation behavior. In this work, we present ViP-Rig, a visual-prompted framework that supports both prompt-first rigging and result-guided editing by injecting features extracted from user-drawn or edited 2D skeletal and rigidity prompts into frozen pretrained backbones. Specifically, ViP-Rig consists of two stages, \textit{Skeleton Generation} and \textit{Skinning Prediction}. In the first stage, the skeletal sketch is processed by the \textit{Dense-to-Compact Visual Prompt Encoding} to produce compact, fixed-length conditioning tokens. The resulting tokens are injected into a frozen pretrained autoregressive generator through gated adapters to control joint placement and branching structure while preserving the generator's geometric prior. In the second stage, the rigidity map is processed using the same visual encoding design, while the pretrained skinning backbone remains frozen. The resulting tokens are symmetrically injected into the point and joint streams to modulate point--joint compatibility and the resulting skinning weights. Experiments on Articulation-XL2.0 and zero-shot evaluation on ModelsResource show that ViP-Rig more accurately recovers target skeletons and skinning weights than geometry-conditioned baselines under prompt-guided evaluation. Qualitative results further demonstrate explicit and localized control in both prompt-first rigging and result-guided editing.
\end{abstract}

\begin{figure*}[!t]
    \centering
    \includegraphics[width=\textwidth]{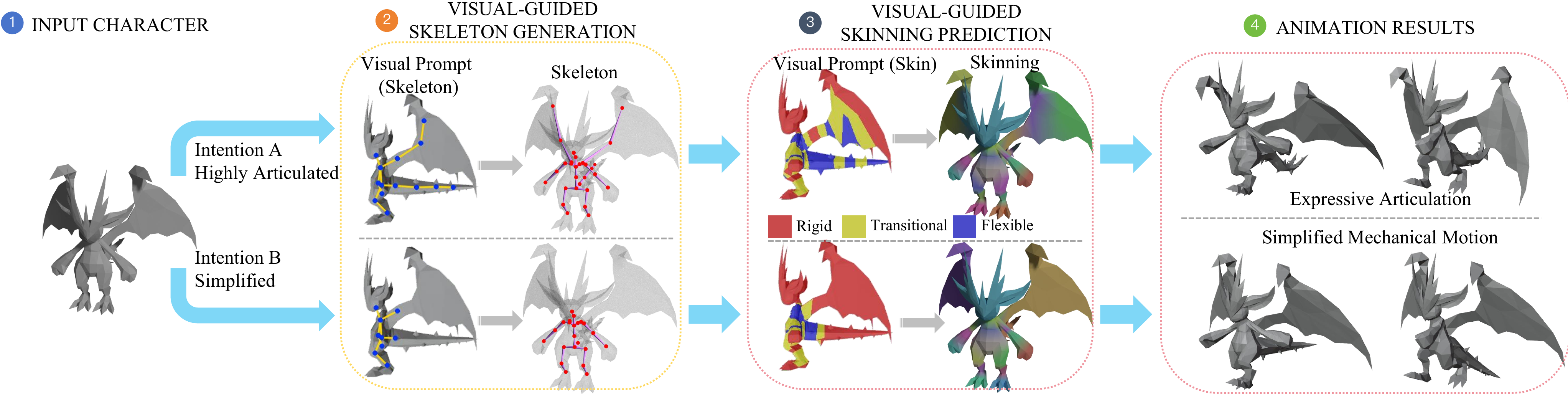}
    \caption{
    \textbf{Overview of ViP-Rig.}
    Given an input mesh, a 2D skeletal sketch controls joint layout and
    branching structure, while a color-coded rigidity map controls regional
    deformation behavior, enabling different task-specific rigs for the same
    geometry.
    }
    \label{fig:teaser}
\end{figure*}


\section{Introduction}

Recent advances in 3D content generation have made high-quality static meshes increasingly accessible, but animation still requires rigging. Rigging comprises skeleton generation, which determines joint locations and hierarchy, and skinning-weight prediction, which assigns each surface point an influence distribution over the joints. Both require substantial expertise and manual effort; artists commonly refine an initial rig until it meets the intended animation requirements. This has motivated extensive work on automatic rigging.

Early learning-based methods predict joints, connectivity, and skinning weights from volumetric or surface geometry~\cite{xu2020rignet,xu2022morig,sun2025drive}. More recent generative approaches model complete hierarchies and point--joint relationships; autoregressive methods serialize skeletons so that joint placement and connectivity are predicted jointly~\cite{song2025magicarticulate,zhang2025unirig,song2025puppeteer}. These advances substantially improve rig completeness and plausibility.

However, these methods seek a plausible rig rather than the particular rig required by a user. Rigging is underdetermined: the same mesh may need a fine-grained or compact hierarchy, with different local joint placements, branch connections, and parent--child relations. Skinning is likewise ambiguous because adjacent regions may require different deformation behavior, such as rigid armor beside a flexible shoulder. UniRig allows users to edit a generated 3D hierarchy before continuing generation~\cite{zhang2025unirig}, but requires professional 3D manipulation. Animator-Centric Skeleton Generation conditions on main bones and global bone density~\cite{sun2026animator}, which cannot specify arbitrary local layouts or branches and does not control regional skinning. An intuitive interface for explicit control of both stages therefore remains missing.

We introduce ViP-Rig, which specifies the desired rig through user-drawn or edited 2D skeletal and rigidity prompts over an aligned mesh rendering. This avoids direct manipulation of a 3D hierarchy or dense per-joint weight painting. Users may draw prompts from scratch for prompt-first rigging, or edit the projected skeleton and derived rigidity map of an existing result; the updated rig can initialize further editing rounds. ViP-Rig then performs \textit{Skeleton Generation} and \textit{Skinning Prediction}. A frozen visual encoder and stage-specific Resampler convert each prompt into compact tokens. Zero-initialized gated adapters inject skeletal tokens into every layer of a frozen autoregressive generator, controlling joint layout and branching while preserving its geometric prior. For skinning, symmetric adapters inject rigidity tokens into the point and joint streams before and after interaction, directly modulating point--joint compatibility.


Our main contributions are summarized as follows:
\begin{itemize}
\item To the best of our knowledge, we present the first visual-prompted framework for controllable rigging of a given mesh, with localized control over skeletal structure and regional skinning in both prompt-first and iterative editing workflows.

\item We introduce \textit{Dense-to-Compact Visual Prompt Encoding} and task-specific injection: layer-wise gated adapters for prior-preserving skeleton control and symmetric adapters for rigidity-aware point--joint matching.

\item Experiments on Articulation-XL2.0 and zero-shot ModelsResource evaluation show improved prompt-guided target recovery, while qualitative results demonstrate direct, localized control in both workflows.
\end{itemize}

\section{Related Work}

\paragraph{Skeleton Generation.}
Early methods construct skeletons through template fitting or medial-axis extraction~\cite{baran2007Pinocchio,yan2018voxel}, but are limited by predefined templates or handcrafted assumptions. Learning-based approaches predict joints from volumetric or surface representations~\cite{xu2020rignet,xu2022morig,sun2025drive}. Autoregressive methods then serialize complete hierarchies to model joint placement and connectivity jointly~\cite{song2025magicarticulate,zhang2025unirig,song2025puppeteer}, with later work adding preference alignment~\cite{guo2025auto} or joint-density control~\cite{sun2026animator}. ViP-Rig instead uses a 2D skeletal prompt for spatially explicit control of joint layout and branching in both generation and iterative editing.

\paragraph{Skinning Weight Prediction.}
Traditional methods derive weights from geometric objectives such as heat diffusion, bounded biharmonic energies, or volumetric geodesics~\cite{baran2007Pinocchio,jacobson2011bounded,dionne2013geodesic}. Learning-based methods learn deformation-aware features from meshes or point clouds~\cite{liu2019neuroskinning,xu2020rignet,pan2021heterskinnet}. Recent approaches use functional diffusion~\cite{song2025magicarticulate,zhang2024functional} or point--joint interaction~\cite{zhang2025unirig,song2025puppeteer}, but remain conditioned mainly on geometry and skeleton structure. These inputs may not distinguish adjacent regions requiring rigid versus smooth deformation. ViP-Rig adds an image-space rigidity prompt to both point and joint representations, enabling explicit regional control and iterative editing.

\section{Method}

\begin{figure*}[t]
    \centering
    \includegraphics[width=\textwidth]{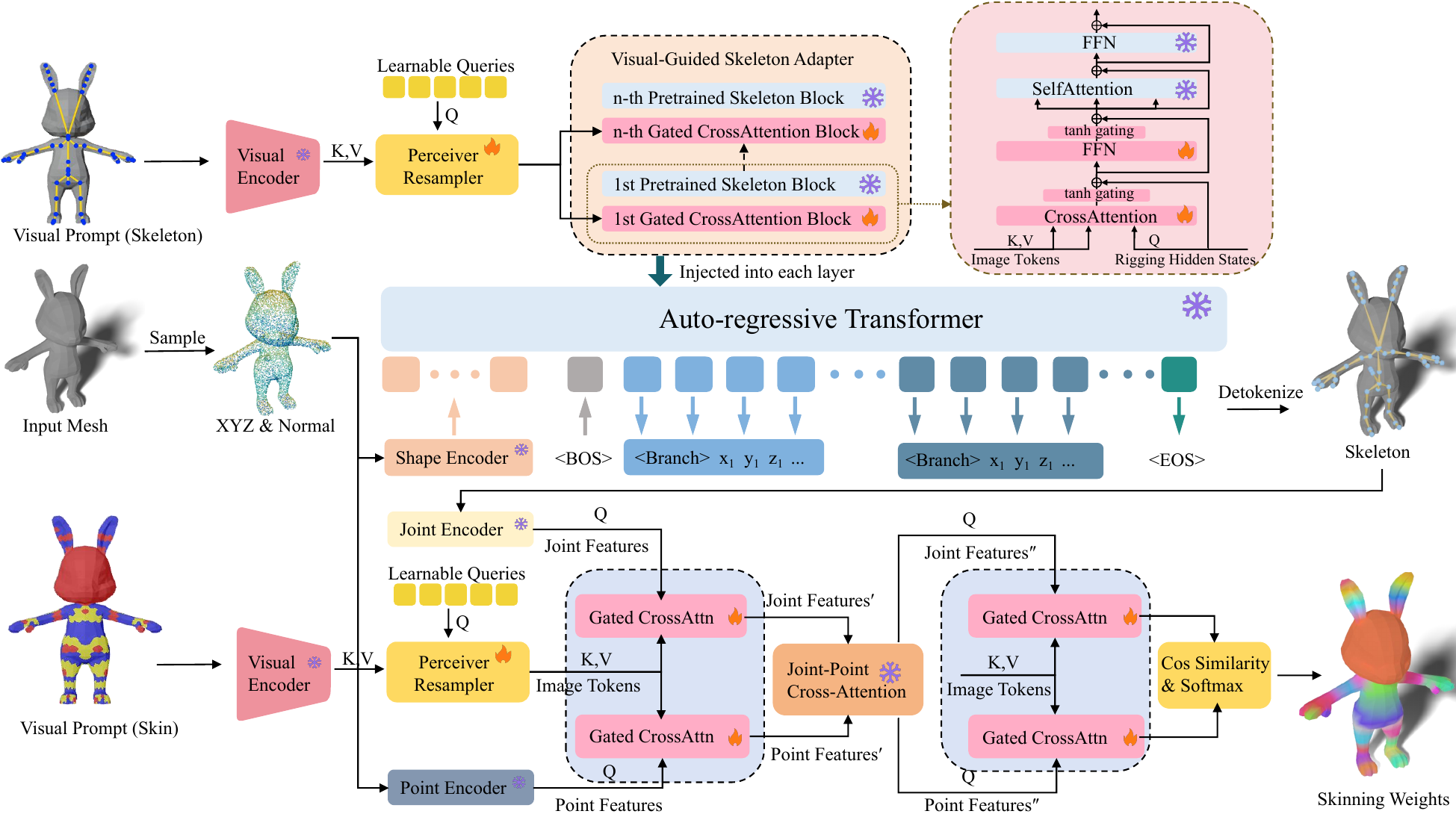}
    \caption{
    \textbf{Architecture of ViP-Rig.}
    Dense visual features are compressed into conditioning tokens and
    injected into frozen pretrained backbones. Layer-wise gated adapters
    guide skeleton generation, while symmetric pre- and post-interaction
    adapters condition point--joint matching for skinning prediction.
    }
    \label{fig:overview}
\end{figure*}

Given an input mesh $\mathcal{M}=(\mathbf{V},\mathbf{F})$, our goal is to generate an animation-ready rig that follows user-specified structural and
deformation requirements, which are represented by two visual prompts defined on a single-view rendering of $\mathcal{M}$. A skeletal prompt $\mathcal{I}_{\mathrm{skel}}$ specifies the desired joint layout and branching structure, while a color-coded rigidity prompt $\mathcal{I}_{\mathrm{skin}}$ marks visible regions as rigid, transitional, or flexible. As illustrated in Figure~\ref{fig:overview}, \textit{ViP-Rig} employs a two-stage framework: \textit{Skeleton Generation} and \textit{Skinning Prediction}. Across both stages, \textit{ViP-Rig} extracts and efficiently injects 2D prompt features while preserving the pretrained geometric priors of the backbones, thereby enabling explicit control over the outputs. The two prompt-conditioned tasks are formulated as:
\begin{equation}
    P\!\left(
        \mathcal{S}
        \mid
        \mathcal{M},
        \mathcal{I}_{\mathrm{skel}}
    \right)
    \quad\text{and}\quad
    P\!\left(
        \mathbf{W}
        \mid
        \mathcal{M},
        \mathcal{S},
        \mathcal{I}_{\mathrm{skin}}
    \right).
    \label{eq:prompt_conditioned_rigging}
\end{equation}
We next describe the two stages in detail.

\subsection{Skeleton Generation}
Following UniRig~\cite{zhang2025unirig}, the skeleton of $\mathcal{M}$ is represented as a graph $\mathcal{S}=(\mathbf{J},\mathcal{P})$, where $\mathbf{J}\in\mathbb{R}^{J\times 3}$ contains the joint positions and $\mathcal{P}$ defines their parent--child hierarchy. UniRig's DFS-based skeleton-tree tokenization serializes $\mathcal{S}$ into a discrete token sequence $\mathbf{s}=(s_1,\ldots,s_T)$ of length $T$, allowing skeleton generation to be formulated as a $T$-step autoregressive process. In this stage, we adopt the pretrained autoregressive Transformer from UniRig~\cite{zhang2025unirig} as the backbone, and use \textit{Dense-to-Compact Visual Prompt Encoding} to encode $\mathcal{I}_{\mathrm{skel}}$ and \textit{Visual-Guided Autoregressive Decoding} to inject the resulting tokens into the Transformer to guide skeleton generation.

\paragraph{Dense-to-Compact Visual Prompt Encoding.}
This module converts $\mathcal{I}_{\mathrm{skel}}$ into compact, fixed-length conditioning tokens for efficient injection into the backbone. Considering that the user-drawn or edited landmarks and connections in $\mathcal{I}_{\mathrm{skel}}$ are defined relative to the visible mesh, their interpretation requires both local prompt cues and the surrounding image context. We therefore use a frozen DINOv2 encoder~\cite{oquab2023dinov2}, denoted by $\mathcal{E}_{\mathrm{vis}}$, to extract dense visual features. However, directly injecting the complete dense token set into the Transformer is unnecessarily costly and retains image content unrelated to skeleton control. We further employ a trainable skeleton-stage Perceiver Resampler~\cite{alayrac2022flamingo}, denoted by $\mathcal{R}_{\mathrm{skel}}$, which uses a fixed set of learnable queries to select prompt-relevant structural cues and compress the dense features into $K$ fixed-length conditioning tokens:
\begin{equation}
    \mathcal{Z}_{\mathrm{skel}}
    =
    \mathcal{R}_{\mathrm{skel}}\!\left(
        \mathcal{E}_{\mathrm{vis}}\!\left(
            \mathcal{I}_{\mathrm{skel}}
        \right)
    \right),
    \qquad
    \mathcal{Z}_{\mathrm{skel}}
    \in
    \mathbb{R}^{K\times d}.
    \label{eq:visual_encoding}
\end{equation}

Because the autoregressive backbone cannot directly process the mesh, we follow the sampling and encoding pipeline of 3DShape2VecSet~\cite{zhang20233dshape2vecset} to convert the input mesh $\mathcal{M}$ into geometry features $\mathbf{F}_{G}$. Once extracted, both $\mathbf{F}_{G}$ and $\mathcal{Z}_{\mathrm{skel}}$ remain fixed throughout the autoregressive decoding process.

\paragraph{Visual-Guided Autoregressive Decoding.}
After obtaining the geometry and visual features, autoregressive decoding proceeds as follows. At step $t$, the backbone uses previous tokens $s_{<t}$ and $\mathbf{F}_{G}$ to predict $s_t$, while prompt tokens $\mathcal{Z}_{\mathrm{skel}}$ are injected into the Transformer to guide generation. To prevent visual injection from overriding the geometry-conditioned prior of the pretrained backbone, we freeze the backbone and insert a \textit{Visual-Guided Skeleton Adapter} into every Transformer layer.

Specifically, let $\mathbf{H}^{(\ell)}\in\mathbb{R}^{L\times d}$ denote the hidden states produced by pretrained layer $\ell$ before visual adaptation,
where $L$ is the current sequence length. The adapter first applies gated cross-attention:
\begin{equation}
    \widetilde{\mathbf{H}}^{(\ell)}
    =
    \mathbf{H}^{(\ell)}
    +
    \tanh(\alpha_{\ell})
    \operatorname{CrossAttn}\!\left(
        \mathbf{H}^{(\ell)},
        \mathcal{Z}_{\mathrm{skel}},
        \mathcal{Z}_{\mathrm{skel}}
    \right),
    \label{eq:skel_cross_attn}
\end{equation}
followed by a gated feed-forward update:
\begin{equation}
    \widehat{\mathbf{H}}^{(\ell)}
    =
    \widetilde{\mathbf{H}}^{(\ell)}
    +
    \tanh(\beta_{\ell})
    \operatorname{FFN}\!\left(
        \widetilde{\mathbf{H}}^{(\ell)}
    \right).
    \label{eq:skel_ffn}
\end{equation}

The layer-specific scalar gates $\alpha_{\ell}$ and $\beta_{\ell}$ are initialized to zero, so each adapter initially performs an identity mapping and preserves the behavior of the pretrained generator. The conditioned states $\widehat{\mathbf{H}}^{(\ell)}$ are then passed to the next pretrained layer, and the final layer predicts the next token. This process is formulated as:
\begin{equation}
    P\!\left(
        \mathbf{s}
        \mid
        \mathcal{M},
        \mathcal{I}_{\mathrm{skel}}
    \right)
    =
    \prod_{t=1}^{T}
    P\!\left(
        s_t
        \mid
        s_{<t},
        \mathbf{F}_{G},
        \mathcal{Z}_{\mathrm{skel}}
    \right).
    \label{eq:skeleton_ar}
\end{equation}
The completed sequence $\mathbf{s}$ is detokenized into joint positions $\mathbf{J}$ and parent--child relations $\mathcal{P}$ to form skeleton $\mathcal{S}$, which is passed to the skinning stage. 

\paragraph{Training.}
For each mesh--skeleton pair $(\mathcal{M},\mathcal{S})$, we render
canonical candidate views and use a vision-language model to select the
view that most clearly presents the ground-truth skeleton with minimal
occlusion.
The joints and parent--child connections are projected onto the selected
mesh rendering and rasterized as 2D landmarks and line segments to form
$\mathcal{I}_{\mathrm{skel}}$.
Geometric and prompt perturbations are then applied to approximate
the sparsity and imprecision of user-provided prompts; detailed rendering
and perturbation settings are provided in the supplementary material.
The pretrained shape encoder, UniRig generator, and DINOv2 encoder remain
frozen, while the skeleton-stage Perceiver Resampler and
Visual-Guided Skeleton Adapters are optimized using the next-token
prediction objective:
\begin{equation}
    \mathcal{L}_{\mathrm{skel}}
    =
    -\sum_{t=1}^{T}
    \log
    P\!\left(
        s_t
        \mid
        s_{<t},
        \mathbf{F}_{G},
        \mathcal{Z}_{\mathrm{skel}}
    \right).
    \label{eq:skel_loss}
\end{equation}

\subsection{Skinning Prediction}

Given the mesh $\mathcal{M}$, skeleton $\mathcal{S}$, and rigidity
prompt $\mathcal{I}_{\mathrm{skin}}$, this stage predicts the
skinning-weight matrix
$\mathbf{W}\in[0,1]^{N\times J}$,
where each row $\mathbf{w}_{i}$ represents the joint-influence
distribution of surface point $i$ and satisfies
$\sum_{j=1}^{J}w_{ij}=1$.
As shown in Figure~\ref{fig:overview},
$\mathcal{I}_{\mathrm{skin}}$ is a color-coded map painted on an aligned
single-view rendering of the mesh, where red, yellow, and blue indicate
rigid, transitional, and flexible deformation behavior, respectively.
Because the prompt specifies only regional deformation tendency rather
than exact per-joint weights, it cannot be directly converted into
$\mathbf{W}$.
We therefore build upon the dual-stream point--joint matching
architecture of Puppeteer~\cite{song2025puppeteer}, which computes a
compatibility score for each point--joint pair and normalizes these
scores over all joints to obtain the corresponding skinning weights.
By injecting the rigidity condition into both feature streams, ViP-Rig
modulates point--joint compatibility and thereby controls the predicted
weight distribution.

\paragraph{Feature preparation.}
We prepare the backbone inputs similarly to the previous stage. Because feature injection is also required, we adopt another \textit{Dense-to-Compact Visual Prompt Encoding} module to extract prompt features $\mathcal{Z}_{\mathrm{skin}}$ from $\mathcal{I}_{\mathrm{skin}}$:
\begin{equation}
    \mathcal{Z}_{\mathrm{skin}}
    =
    \mathcal{R}_{\mathrm{skin}}\!\left(
        \mathcal{E}_{\mathrm{vis}}\!\left(
            \mathcal{I}_{\mathrm{skin}}
        \right)
    \right),
    \qquad
    \mathcal{Z}_{\mathrm{skin}}
    \in
    \mathbb{R}^{K\times d}.
    \label{eq:skin_visual_encoding}
\end{equation}

We then reuse the joint and point encoders of the Puppeteer backbone to encode the skeleton $\mathcal{S}$ and the mesh $\mathcal{M}$, producing joint features $\mathbf{F}_{J}\in\mathbb{R}^{J\times d}$ and surface-point features $\mathbf{F}_{P}\in\mathbb{R}^{N\times d}$, respectively.

\paragraph{Symmetric Feature Injection.}
Similar to the previous stage, we freeze the backbone and inject the prompt feature $\mathcal{Z}_{\mathrm{skin}}$ into both the point and joint streams, allowing user-specified regional behavior to alter point--joint compatibility. We refer to the frozen point--joint interaction block inherited from Puppeteer as the Joint--Point Cross-Attention module (JPCA), which exchanges geometric and skeletal context between the two streams. Because skinning weights are determined by matching point and joint features, we condition both sides of the compatibility function. Injection before $\operatorname{JPCA}$ makes the exchanged point--joint context prompt-aware, while injection after $\operatorname{JPCA}$ directly refines the representations used for final matching.

Specifically, we instantiate four independently parameterized copies of the gated cross-attention--FFN adapter defined in Eqs.~\ref{eq:skel_cross_attn}--\ref{eq:skel_ffn}:
$\mathcal{G}_{P}^{\mathrm{pre}}$,  $\mathcal{G}_{J}^{\mathrm{pre}}$, $\mathcal{G}_{P}^{\mathrm{post}}$, and $\mathcal{G}_{J}^{\mathrm{post}}$. Each adapter has separate parameters and zero-initialized scalar gates. Before point--joint interaction, the initial point and joint features are conditioned independently:
\begin{equation}
\begin{aligned}
    \mathbf{F}_{P}^{\mathrm{pre}}
    &=
    \mathcal{G}_{P}^{\mathrm{pre}}\!\left(
        \mathbf{F}_{P},
        \mathcal{Z}_{\mathrm{skin}}
    \right),\\
    \mathbf{F}_{J}^{\mathrm{pre}}
    &=
    \mathcal{G}_{J}^{\mathrm{pre}}\!\left(
        \mathbf{F}_{J},
        \mathcal{Z}_{\mathrm{skin}}
    \right).
\end{aligned}
\label{eq:skinning_pre_injection}
\end{equation}
The prompt-aware representations are passed to the frozen Joint--Point Cross-Attention module:
\begin{equation}
    \left(
        \overline{\mathbf{F}}_{P},
        \overline{\mathbf{F}}_{J}
    \right)
    =
    \operatorname{JPCA}\!\left(
        \mathbf{F}_{P}^{\mathrm{pre}},
        \mathbf{F}_{J}^{\mathrm{pre}}
    \right).
    \label{eq:skinning_jpca}
\end{equation}
Here, $\overline{\mathbf{F}}_{P}$ contains surface representations informed by the skeleton, while $\overline{\mathbf{F}}_{J}$ contains joint representations informed by the mesh surface. The exchanged features are conditioned again immediately before final matching:
\begin{equation}
\begin{aligned}
    \widehat{\mathbf{F}}_{P}
    &=
    \mathcal{G}_{P}^{\mathrm{post}}\!\left(
        \overline{\mathbf{F}}_{P},
        \mathcal{Z}_{\mathrm{skin}}
    \right),\\
    \widehat{\mathbf{F}}_{J}
    &=
    \mathcal{G}_{J}^{\mathrm{post}}\!\left(
        \overline{\mathbf{F}}_{J},
        \mathcal{Z}_{\mathrm{skin}}
    \right).
\end{aligned}
\label{eq:skinning_post_injection}
\end{equation}
Together, the injections allow the rigidity prompt to modify both sides of point--joint matching.

Finally, following Puppeteer~\cite{song2025puppeteer}, we compute
skinning weights by applying a joint-wise softmax to the cosine
similarities between the final point and joint features:
\begin{equation}
    w_{ij}
    =
    \operatorname{softmax}_{j}\!\left(
        \gamma\,
        \frac{
            \widehat{\mathbf{f}}_{P,i}^{\top}
            \widehat{\mathbf{f}}_{J,j}
        }{
            \|\widehat{\mathbf{f}}_{P,i}\|_{2}
            \|\widehat{\mathbf{f}}_{J,j}\|_{2}
        }
    \right),
    \label{eq:skinning_softmax}
\end{equation}
where $\gamma$ is the frozen similarity scale inherited from the
pretrained matching operation.

\paragraph{Training.}
We train the skinning stage independently with ground-truth skeletons so
that weight prediction is not affected by errors from the preceding
skeleton stage.
Existing rigging datasets provide ground-truth skinning weights but no
rigidity prompts, so we derive the training prompt from the concentration
of each ground-truth joint-influence distribution.
Let $\mathbf{q}_{i}=(q_{i1},\ldots,q_{iJ})\in[0,1]^{J}$ denote the
ground-truth weights of point $i$, with
$\sum_{j=1}^{J}q_{ij}=1$.
We compute its normalized Shannon entropy
\begin{equation}
    \overline{H}_i
    =
    -\frac{1}{\log J}
    \sum_{j=1}^{J}
    q_{ij}\log\!\left(q_{ij}+\epsilon\right),
    \label{eq:skinning_entropy}
\end{equation}
where $\epsilon>0$ ensures numerical stability, and normalization by
$\log J$ makes the measure comparable across skeletons with different
numbers of joints.
Low and high entropy indicate concentrated and distributed joint
influences, respectively, which we use as proxies for rigid and flexible
behavior.
We discretize $\overline{H}_i$ into three rigidity levels and render the
resulting field on the selected mesh view to obtain
$\mathcal{I}_{\mathrm{skin}}$.
The prompt encodes only regional influence concentration rather than the
identities or exact weights of the influencing joints.
We use the same VLM-selected view as in skeleton generation and apply
conservative perturbations that preserve point--joint correspondence;
further details are provided in the supplementary material.

The pretrained point and joint encoders, Joint--Point Cross-Attention
module, matching operation, and DINOv2 encoder remain frozen.
Only the skinning-stage Perceiver Resampler and four visual-guided
adapters are optimized.
We minimize the KL divergence between the target and predicted
joint-influence distributions:
\begin{equation}
    \mathcal{L}_{\mathrm{skin}}
    =
    \sum_{i=1}^{N}
    D_{\mathrm{KL}}\!\left(
        \mathbf{q}_{i}
        \,\middle\|\,
        \mathbf{w}_{i}
    \right).
    \label{eq:skin_loss}
\end{equation}

\section{Experiments}

\subsection{Experimental Setup}

\paragraph{Datasets.}
We train and evaluate ViP-Rig on Articulation-XL2.0
(Art-XL2.0)~\cite{song2025puppeteer}, using 46K high-quality training
samples and a held-out test set of 2K samples.
To evaluate cross-dataset generalization, we additionally report zero-shot
results on the 270 test characters of
ModelsResource~\cite{xu2019predicting}.
No ModelsResource samples are used for training or model selection.

\paragraph{Data processing and training.}
Following prior protocols~\cite{song2025puppeteer,zhang2025unirig},
mesh surfaces are sampled and normalized to canonical coordinates.
Detailed sampling and data-augmentation settings for both stages are
provided in the supplementary material.
Both stages are trained independently for 100 epochs with AdamW, a
one-cycle schedule peaking at $1\times10^{-4}$, bf16 precision, and a
per-GPU batch size of 64.
The skeleton and skinning stages use eight and two NVIDIA H20 GPUs,
respectively; all pretrained backbones and the DINOv2 encoder remain frozen.

\paragraph{Evaluation protocol.}
We evaluate ViP-Rig from two complementary perspectives.
For quantitative evaluation, benchmark prompts are constructed from
ground-truth annotations using the same projection and rendering
procedures employed during training.
They specify which valid rig is intended for an otherwise ambiguous
input geometry, so the benchmark measures prompt-guided target recovery.
Because existing baselines do not accept visual prompts, they retain
their standard inputs and serve as geometry-conditioned references.
Skeleton metrics use predicted skeletons, while all skinning methods use
ground-truth skeletons to isolate weight quality.
For qualitative evaluation of controllability, we additionally use prompts
drawn from scratch and prompts obtained by editing existing predictions,
corresponding to prompt-first rigging and result-guided editing,
respectively. These prompts are not derived from benchmark annotations.

\begin{figure*}[!t]
    \centering
    \includegraphics[
        width=\textwidth,
        height=0.58\textheight,
        keepaspectratio
    ]{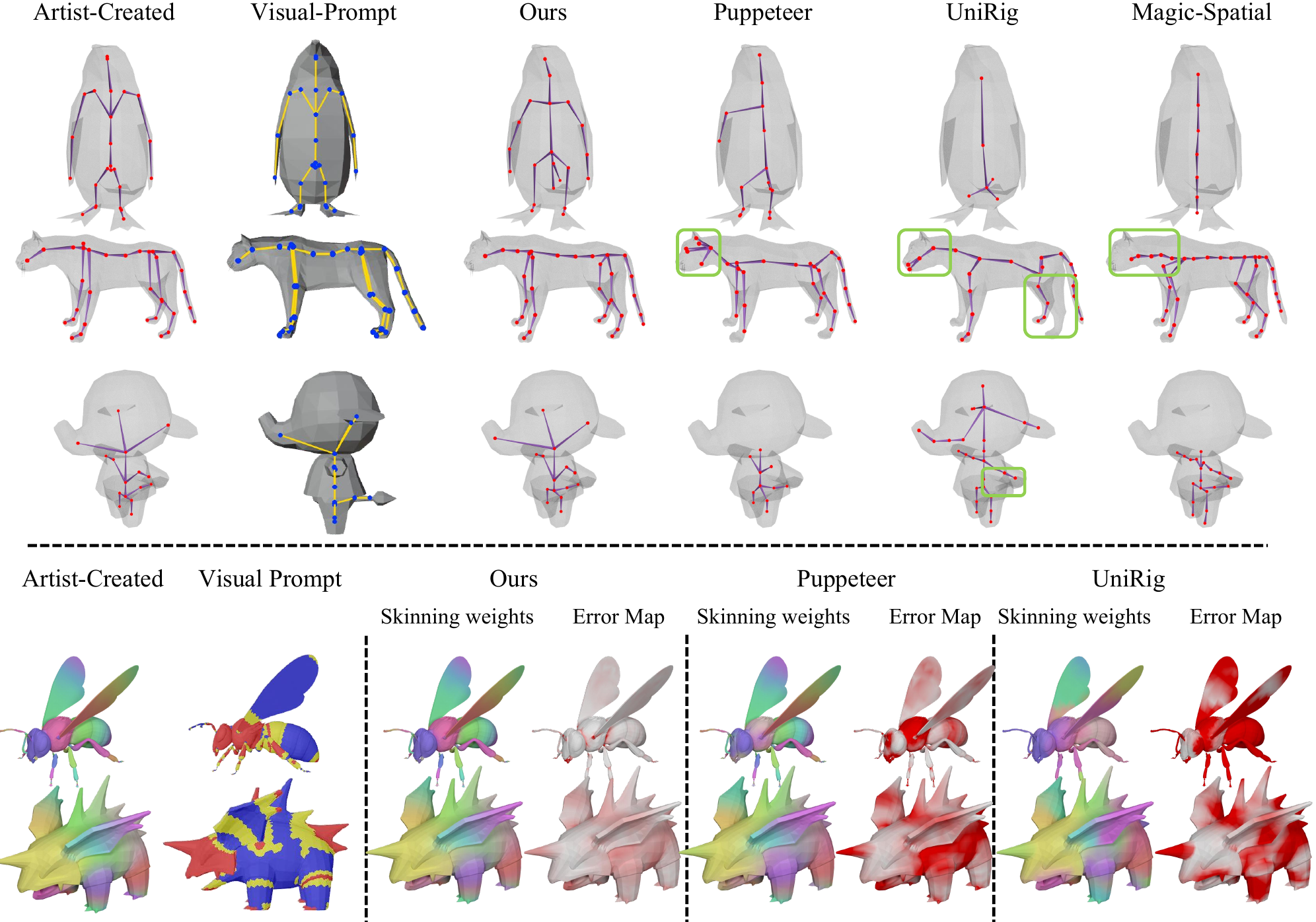}

    \caption{
    \textbf{Qualitative comparisons under the benchmark protocol.}
    \textbf{Top:} ViP-Rig better follows prompt-specified joint layouts and
    branches than geometry-conditioned baselines. \textbf{Bottom:}
    Skinning-weight predictions and error maps show that rigidity prompts
    improve local influence assignment in adjacent regions with different
    deformation behavior.
    }
    \label{fig:qualitative_results}
\end{figure*}

\subsection{Skeleton Generation Results}

\paragraph{Baselines and metrics.}
We compare ViP-Rig with RigNet~\cite{xu2020rignet},
the hierarchical and spatial serialization variants of
MagicArticulate~\cite{song2025magicarticulate},
UniRig~\cite{zhang2025unirig}, and
Puppeteer~\cite{song2025puppeteer}.
The main paper reports joint-to-joint, joint-to-bone, and bone-to-bone
Chamfer distances (CD-J2J, CD-J2B, and CD-B2B).
CD-J2J measures joint-location alignment, whereas CD-J2B and CD-B2B are
less sensitive to different joint sampling densities and evaluate the
agreement of the underlying skeletal segments.
All distances are normalized by object scale and reported as percentages.

\paragraph{Quantitative and qualitative comparison.}
Table~\ref{tab:skeleton_results} shows that ViP-Rig obtains the lowest
three Chamfer distances on both Art-XL2.0 and ModelsResource.
The qualitative comparisons at the top of
Figure~\ref{fig:qualitative_results} clarify where these improvements
arise.
In locally complex or geometrically ambiguous regions,
geometry-conditioned baselines tend to simplify fine and peripheral
structures, omit target joints and branches, or introduce spurious bones
unrelated to the intended structure.
In contrast, ViP-Rig follows the prompt-specified landmarks and
connections while using the pretrained geometric prior to complete
plausible 3D depths and hierarchies.
Its consistent improvement on ModelsResource further indicates that the
visual prompt remains effective in resolving target-specific skeletal
ambiguity under zero-shot cross-dataset transfer.

\begin{table}[t]
\centering
\small
\setlength{\tabcolsep}{3.2pt}
\caption{
Skeleton-generation results.
Lower is better; MR denotes ModelsResource.
}
\label{tab:skeleton_results}
\begin{tabular}{@{}llccc@{}}
\toprule
Data & Method
& J2J $\downarrow$
& J2B $\downarrow$
& B2B $\downarrow$ \\
\midrule
Art
& RigNet        & 7.587\% & 6.347\% & 6.366\% \\
& Magic-hier    & 3.435\% & 2.757\% & 2.393\% \\
& UniRig        & 3.232\% & 2.540\% & 2.124\% \\
& Magic-spatial & 3.041\% & 2.479\% & 2.099\% \\
& Puppeteer     & 2.926\% & 2.193\% & 1.841\% \\
& Ours          & \textbf{2.442\%}
                & \textbf{1.825\%}
                & \textbf{1.558\%} \\
\midrule
MR
& RigNet        & 6.375\% & 5.115\% & 5.245\% \\
& Magic-hier    & 4.119\% & 3.155\% & 2.780\% \\
& UniRig        & 3.797\% & 2.888\% & 2.437\% \\
& Magic-spatial & 3.920\% & 3.138\% & 2.712\% \\
& Puppeteer     & 3.760\% & 2.745\% & 2.189\% \\
& Ours          & \textbf{3.281\%}
                & \textbf{2.312\%}
                & \textbf{1.916\%} \\
\bottomrule
\end{tabular}
\end{table}

\subsection{Skinning Weight Prediction Results}

\paragraph{Baselines and metrics.}
We compare ViP-Rig with Geodesic Voxel Binding
(GVB)~\cite{dionne2013geodesic},
RigNet~\cite{xu2020rignet},
MagicArticulate~\cite{song2025magicarticulate},
UniRig~\cite{zhang2025unirig}, and
Puppeteer~\cite{song2025puppeteer}.
Every method is evaluated with the ground-truth skeleton.
Following Puppeteer, an influence is considered nonzero when its weight
exceeds $10^{-4}$.
Precision and recall measure the overlap between predicted and
ground-truth active joint influences, while $\ell_1$ error measures the
average absolute difference of the complete weight vectors over all
surface points.

\paragraph{Quantitative and qualitative comparison.}
Table~\ref{tab:skinning_results} shows that ViP-Rig consistently
outperforms all baselines across precision, recall, and $\ell_1$ error
on both datasets.
The weight visualizations and error maps at the bottom of
Figure~\ref{fig:qualitative_results} show how these improvements
translate into more accurate local influence assignments.
Without explicit regional deformation cues, geometry-conditioned
methods may diffuse weights across nearby rigid or functionally
unrelated parts.
The rigidity prompt supplies complementary region-level deformation
cues, allowing ViP-Rig to distinguish geometrically adjacent regions
that should exhibit different behavior.
Its predictions consequently align more closely with the artist-created
references and exhibit weaker, more localized errors.
The consistent improvement on ModelsResource further indicates that
these regional cues remain effective under zero-shot cross-dataset
transfer.

\begin{table}[t]
\centering
\small
\setlength{\tabcolsep}{2.7pt}
\caption{
Skinning-weight prediction.
Higher precision and recall are better; lower $\ell_1$ is better.
}
\label{tab:skinning_results}
\begin{tabular}{@{}lrrrrrr@{}}
\toprule
& \multicolumn{3}{c}{Art-XL2.0}
& \multicolumn{3}{c}{MR} \\
\cmidrule(lr){2-4}
\cmidrule(lr){5-7}
Method
& P. $\uparrow$ & R. $\uparrow$ & $\ell_1$ $\downarrow$
& P. $\uparrow$ & R. $\uparrow$ & $\ell_1$ $\downarrow$ \\
\midrule
GVB        & 72.9\% & 65.5\% & 0.745 & 69.3\% & 79.2\% & 0.687 \\
RigNet     & 73.7\% & 66.1\% & 0.729 & 65.7\% & 80.2\% & 0.707 \\
MagicArti. & 74.6\% & 71.3\% & 0.451 & 68.1\% & 80.7\% & 0.642 \\
UniRig     & 78.7\% & 52.9\% & 0.790 & 72.3\% & 68.6\% & 0.703 \\
Puppeteer  & 87.6\% & 74.0\% & 0.335 & 79.7\% & 81.6\% & 0.443 \\
Ours       & \textbf{88.3\%} & \textbf{76.6\%} & \textbf{0.309}
           & \textbf{80.9\%} & \textbf{81.9\%} & \textbf{0.421} \\
\bottomrule
\end{tabular}
\end{table}

\begin{figure*}[!t]
    \centering
    \begin{minipage}[t]{0.492\textwidth}
        \centering
        \includegraphics[
            width=\linewidth,
            height=0.255\textheight,
            keepaspectratio
        ]{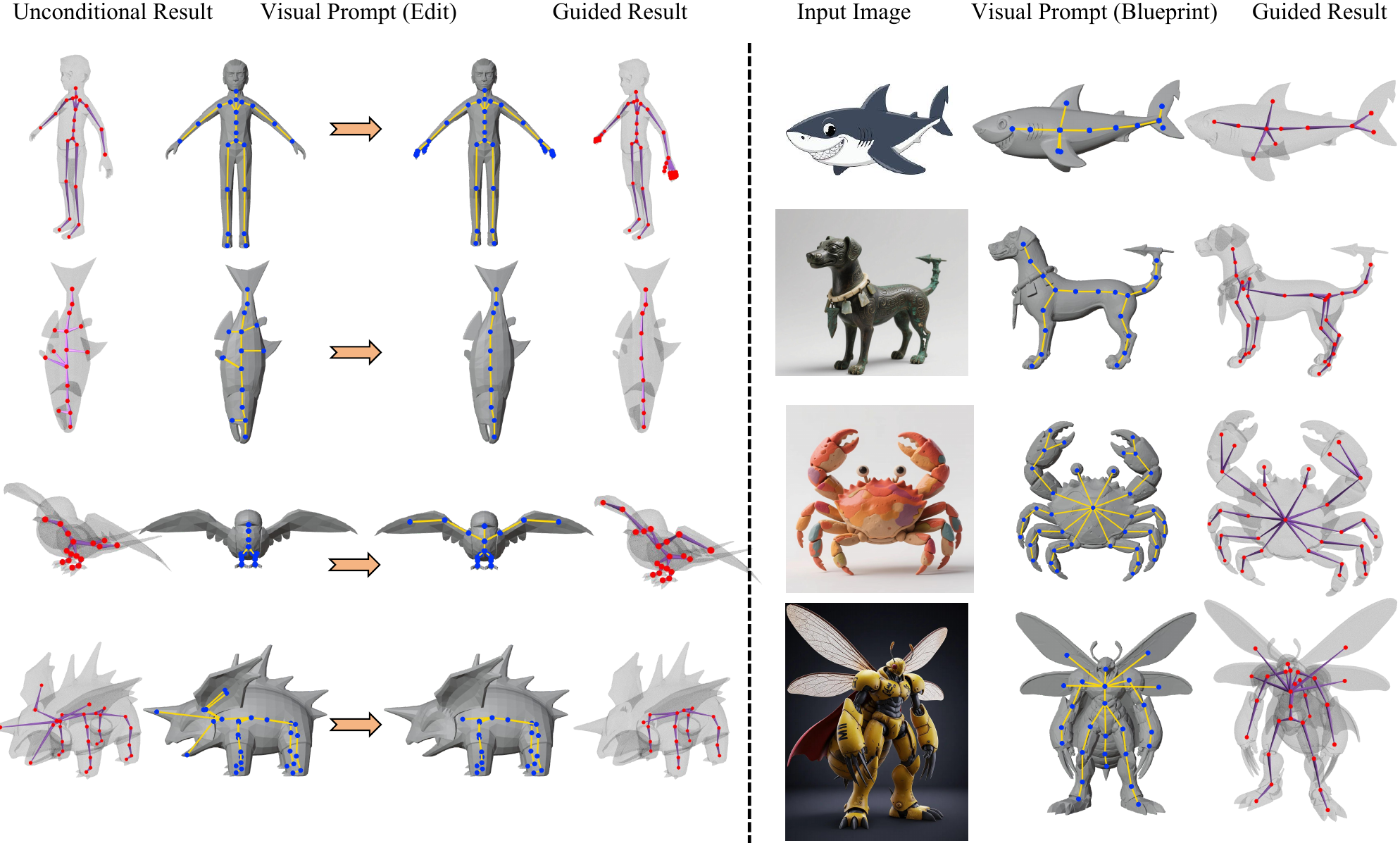}
    \end{minipage}
    \hfill
    \begin{minipage}[t]{0.492\textwidth}
        \centering
        \includegraphics[
            width=\linewidth,
            height=0.255\textheight,
            keepaspectratio
        ]{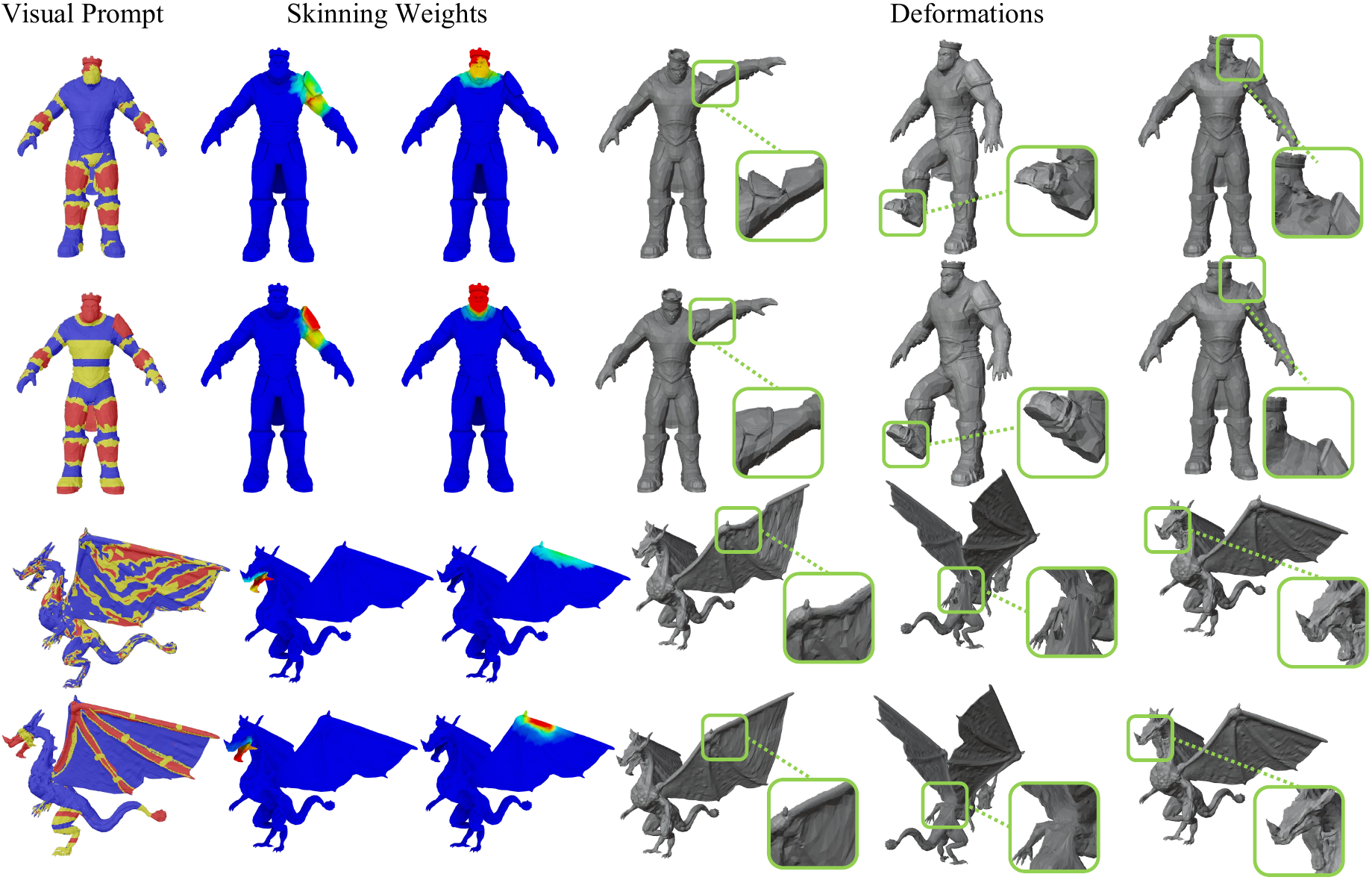}
    \end{minipage}
    \caption{
    \textbf{Controllable rigging applications.}
    \textbf{Left:} Users edit the projected skeleton of an unconditional
    UniRig result obtained with the visual adapters disabled to elaborate or
    simplify its structure, or provide a skeletal blueprint to rig an
    unrigged mesh directly, including meshes generated by an off-the-shelf
    image-to-3D system.
    \textbf{Right:} Users edit a rigidity map derived from an unconditional
    Puppeteer prediction to modify local skinning weights and deformation
    behavior; green insets highlight local details.
    }
    \label{fig:applications}
\end{figure*}

\subsection{Qualitative Evaluation of Controllability}

Figure~\ref{fig:applications} complements the annotation-derived
quantitative benchmark with prompts drawn from scratch and prompts obtained
by editing existing predictions. These examples evaluate whether ViP-Rig follows
user-specified changes beyond recovering benchmark targets, covering both
prompt-first rigging and result-guided editing.

\paragraph{Skeleton editing and prompt-first rigging.}
To assess whether ViP-Rig supports both direct specification and iterative
editing of skeletal structures, we evaluate result-guided editing and
prompt-first rigging in the left part of Figure~\ref{fig:applications}.
For result-guided editing, the frozen UniRig backbone first produces an
unconditional skeleton with the visual adapters disabled. Users edit its
projected structure to add joints and branches for a more detailed
hierarchy or remove unnecessary structures for a simpler rig, and the
edited projection is then used as the skeletal prompt for ViP-Rig. The
resulting skeleton can be projected again to initialize subsequent editing
rounds.
Alternatively, prompt-first rigging does not require an initial skeleton:
users draw a skeletal blueprint directly over the rendering of an unrigged
mesh, and ViP-Rig generates a corresponding 3D skeleton by combining the
prompt with the mesh geometry. The examples include meshes produced by an
off-the-shelf image-to-3D system, illustrating the applicability of ViP-Rig
to diverse AI-generated assets.

\paragraph{Localized rigidity control.}
To assess whether local rigidity edits produce corresponding changes in
skinning weights and deformation behavior, we perform result-guided editing
on selected regions in the right part of Figure~\ref{fig:applications}.
We first obtain an unconditional skinning prediction from the frozen
Puppeteer backbone with the visual adapters disabled and convert its
joint-influence distributions into an initial rigidity map using the same
normalized-entropy mapping as in training. Users edit selected regions of
this map to request greater rigidity or flexibility, and the edited map is
used as the rigidity prompt for ViP-Rig. The updated skinning weights produce corresponding changes in local
deformation during animation and can be converted into a new rigidity map
for subsequent editing rounds.

\subsection{Ablation Studies}

\paragraph{Skeleton visual fusion.}
To verify whether skeletal prompts require persistent guidance throughout
autoregressive decoding, Table~\ref{tab:skeleton_ablation} compares
visual-token prefixing with zero-initialized gated adapters at different
insertion frequencies.
Prefixing performs worse than the gated adapters, and accuracy improves
consistently as the adapters are inserted more frequently, from CA-4 to
Full.
This supports persistent visual guidance throughout autoregressive
coordinate and branch prediction.

\begin{table}[t]
\centering
\footnotesize
\setlength{\tabcolsep}{1.8pt}
\caption{
Skeleton-fusion ablation.
CA-$k$ denotes cross-attention every $k$ layers; lower is better.
}
\label{tab:skeleton_ablation}
\begin{tabular}{@{}lrrrrrr@{}}
\toprule
& \multicolumn{3}{c}{Art-XL2.0}
& \multicolumn{3}{c}{MR} \\
\cmidrule(lr){2-4}
\cmidrule(lr){5-7}
Variant
& J2J $\downarrow$ & J2B $\downarrow$ & B2B $\downarrow$
& J2J $\downarrow$ & J2B $\downarrow$ & B2B $\downarrow$ \\
\midrule
Prefix
& 3.521\% & 2.813\% & 2.596\%
& 3.825\% & 3.020\% & 2.571\% \\
CA-4
& 3.017\% & 2.154\% & 1.901\%
& 3.541\% & 2.601\% & 2.112\% \\
CA-2
& 2.653\% & 1.915\% & 1.672\%
& 3.401\% & 2.392\% & 2.053\% \\
Full
& \textbf{2.442\%} & \textbf{1.825\%} & \textbf{1.558\%}
& \textbf{3.281\%} & \textbf{2.312\%} & \textbf{1.916\%} \\
\bottomrule
\end{tabular}
\end{table}

\paragraph{Rigidity-aware skinning injection.}
To verify whether rigidity control should modify point--joint compatibility
in feature space and condition both matching streams,
Table~\ref{tab:skinning_ablation} compares dual-path injection with an
explicit similarity-scale baseline and single-path variants.
Such scalar modulation can only sharpen or flatten a distribution, whereas
feature-space injection can change point--joint compatibility.
Explicit modulation is therefore weaker, and removing either the point
or joint pathway also degrades all metrics.
The full model performs best on both datasets, supporting symmetric
conditioning of both feature streams.

\begin{table}[t]
\centering
\small
\setlength{\tabcolsep}{2.7pt}
\caption{
Skinning-injection ablation.
Higher precision and recall are better; lower $\ell_1$ is better.
}
\label{tab:skinning_ablation}
\begin{tabular}{@{}lrrrrrr@{}}
\toprule
& \multicolumn{3}{c}{Art-XL2.0}
& \multicolumn{3}{c}{MR} \\
\cmidrule(lr){2-4}
\cmidrule(lr){5-7}
Variant
& P. $\uparrow$ & R. $\uparrow$ & $\ell_1$ $\downarrow$
& P. $\uparrow$ & R. $\uparrow$ & $\ell_1$ $\downarrow$ \\
\midrule
Explicit
& 86.1\% & 76.0\% & 0.331
& 79.4\% & 81.6\% & 0.439 \\
w/o point
& 86.9\% & 75.8\% & 0.323
& 80.0\% & 81.5\% & 0.435 \\
w/o joint
& 87.5\% & 76.1\% & 0.314
& 80.3\% & 81.4\% & 0.430 \\
Full
& \textbf{88.3\%} & \textbf{76.6\%} & \textbf{0.309}
& \textbf{80.9\%} & \textbf{81.9\%} & \textbf{0.421} \\
\bottomrule
\end{tabular}
\end{table}

\section{Conclusion}

ViP-Rig conditions frozen skeleton and skinning backbones on complementary
structural and rigidity prompts through layer-wise gated adapters and
symmetric pre- and post-interaction adapters.
Under prompt-guided evaluation, it consistently improves target recovery
over geometry-conditioned baselines on Art-XL2.0 and under zero-shot
evaluation on ModelsResource.
Qualitative results further demonstrate localized control over skeletal
topology and deformation behavior in both prompt-first rigging and
result-guided editing.

\bibliography{aaai2027}


\end{document}